\documentclass[letterpaper, 10 pt, conference]{ieeeconf}  %

\IEEEoverridecommandlockouts                              %

\usepackage[T1]{fontenc}     %
\usepackage[utf8]{inputenc}
\usepackage{graphicx} %
\usepackage{mathptmx} %
\usepackage{times} %
\usepackage{pifont}
\newcommand{\cmark}{\ding{51}}%
\newcommand{\xmark}{\ding{55}}%
\usepackage{amsmath} %
\usepackage{amssymb}  %
\usepackage{url}

\usepackage[x11names,rgb,usenames,dvipsnames]{xcolor}
\usepackage{tikz}

\usepackage{booktabs, tabularx, multirow}
\usepackage{ragged2e} %
\usepackage{units}    %
\usepackage{nicefrac} %

\usepackage{caption}
\usepackage{subcaption}

\usepackage{paralist}
\usepackage{multicol}

\title{\LARGE \bf
YCB-M: A Multi-Camera RGB-D Dataset for Object Recognition and 6DoF Pose Estimation
}

\author{Till Grenzdörffer$^{*,1,2}$, Martin Günther$^{*,1}$ and Joachim Hertzberg$^{1,2}$%
\thanks{This work is supported by the CoPDA project through a grant of the German Federal Ministry of Education and Research (BMBF) with Grant Number 01IW19003. The DFKI Niedersachsen Lab (DFKI NI) is sponsored by the Ministry of Science and Culture of Lower Saxony and the VolkswagenStiftung.}%
\thanks{$^{*}$ These authors contributed equally to this work.}%
\thanks{$^{1}$ German Research Center for Artificial Intelligence (DFKI), 49090 Osnabrück, Germany
{\tt \{till.grenzdoerffer, martin.guenther, joachim.hertzberg\}@dfki.de}}%
\thanks{$^{2}$ Osnabrück University, Institute of Computer Science, Wachsbleiche 27, 49090 Osnabrück, Germany}%
}

\begin{document}

\onecolumn

\noindent
© 2020 IEEE. Personal use of this material is permitted. Permission from IEEE
must be obtained for all other uses, in any current or future media, including
reprinting/republishing this material for advertising or promotional purposes,
creating new collective works, for resale or redistribution to servers or
lists, or reuse of any copyrighted component of this work in other works. \\

\noindent
\textbf{Full citation:}\\

\begin{quote}
T.~Grenzdörffer, M.~Günther, and J.~Hertzberg, ``{YCB-M}: A Multi-Camera
  {RGB-D} Dataset for Object Recognition and {6DoF} Pose Estimation,'' in
  \emph{2020 {IEEE} International Conference on Robotics and Automation ({ICRA}),
  Paris, France, May 31--June 4, 2020}. \hskip 1.1em\relax {IEEE}, 2020, pp.
  3650--3656. doi: 10.1109/ICRA40945.2020.9197426.
\end{quote}\mbox{}\\

\noindent
Final, published article: \textcolor{blue}{\url{https://ieeexplore.ieee.org/document/9197426}}

\twocolumn
\newpage

\maketitle
\thispagestyle{empty}
\pagestyle{empty}

\begin{abstract}

  While a great variety of 3D cameras have been introduced in recent years,
  most publicly available datasets for object recognition and pose estimation
  focus on one single camera.
  In this work, we present a dataset of 32 scenes that have been captured by 7
  different 3D cameras, totaling 49,294 frames. This allows evaluating the
  sensitivity of pose estimation algorithms to the specifics of the used camera
  and the development of more robust algorithms that are more independent of
  the camera model. Vice versa, our dataset enables researchers to perform a
  quantitative comparison of the data from several different cameras and depth
  sensing technologies and evaluate their algorithms before selecting a camera
  for their specific task.
  The scenes in our dataset contain 20 different objects from the common
  benchmark YCB object and model set \cite{Calli2015icar,Calli2017ijrr}. We
  provide full ground truth 6DoF poses for each object, per-pixel segmentation,
  2D and 3D bounding boxes and a measure of the amount of occlusion of each
  object.
  We have also performed an initial evaluation of the cameras using our dataset
  on a state-of-the-art object recognition and pose estimation system
  \cite{Tremblay2018dope}.
\end{abstract}

\section{Introduction}

Since the advent of the Microsoft Kinect in 2010, 3D cameras have become the
predominant sensor for many robotic tasks, ranging from object detection,
semantic segmentation and scene understanding to
grasping and 3D mapping. However, this first generation of Primesense-based
RGB-D cameras like the Kinect and ASUS Xtion has reached its end-of-life and is
no longer available. In recent years, a plethora of 3D cameras have become
available to take their place. These cameras differ in many aspects such as the sensing technology
(Structured Light, Time-of-Flight, (active) stereo), range, field of view, size
and power consumption. At the same time, different robots and robotic tasks
have different requirements; while a mobile manipulator might require a
high-resolution RGB-D camera with good depth resolution, a small AUV requires a
small and light-weight camera with low power consumption (as long as the data
quality is sufficient). However, it is difficult to choose the correct camera
based on the manufacturer's specifications alone; on the other hand, buying and
setting up different cameras is expensive and labor-intensive.

For this reason, we present the YCB-Multicam (YCB-M) dataset, which includes
RGB and depth data from seven different 3D cameras.  One popular task that
required for a variety of robotic applications (such as object manipulation,
semantic mapping or human-robot communication) is the recognition and 6DoF pose
estimation of objects. Therefore, our dataset focuses on this task and includes
6DoF ground truth pose annotations for each object in each of the 49,294 frames
in the dataset. By providing this dataset to the public, we hope to support
research in object recognition and pose estimation as well as enable
researchers to evaluate different cameras for related settings beyond pose
estimation.

In the remainder of this paper, we first discuss related work, describe our
data acquisition and ground truth labeling methodology, give an overview of
the used 3D cameras and finally present an initial evaluation of a
state-of-the-art object recognition and pose estimation algorithm on our
dataset.

\section{Related Work}

We have grouped the related work in two parts: datasets and comparisons of
multiple 3D cameras.

\subsection{Datasets}

To place our dataset in the context of existing datasets and clarify our
contribution, we have summarized related datasets in
Table~\ref{tab:dataset_comparison}.

\begin{table*}
  \caption{Overview of related datasets}
  \label{tab:dataset_comparison}

  \begin{tabularx}{\linewidth}{llXXXXXXXXXXX}
    \toprule
    & \rotatebox{90}{purpose} & \rotatebox{90}{data source} & \rotatebox{90}{\# object classes} & \rotatebox{90}{\# frames} & \rotatebox{90}{\# scenes} & \rotatebox{90}{\# simult.\ cams} & \rotatebox{90}{depth} & \rotatebox{90}{6DoF poses} & \rotatebox{90}{segmentation} & \rotatebox{90}{2d bbox} & \rotatebox{90}{3d bbox} & \rotatebox{90}{mesh models} \\ \midrule
    YCB-M \textbf{(ours)}                 & pose estimation       & real        & 20                & 49k       & 32        & 7               & \cmark& \cmark     & \cmark       & \cmark  & \cmark  & \cmark      \\
    YCB-Video \cite{Xiang2017corr}        & pose estimation       & real        & 21                & 134k      & 92        & 1               & \cmark& \cmark     & \cmark       & \cmark  & \cmark  & \cmark      \\
    FAT \cite{Tremblay2018fat}            & pose estimation       & synthetic   & 21                & 60k       & 3075      & 1               & \cmark& \cmark     & \cmark       & \cmark  & \cmark  & \cmark      \\
    LINEMOD \cite{Hinterstoisser2012accv} & pose estimation       & real        & 15                & 18k       & 15        & 1               & \cmark& \cmark$^1$ & \xmark       & \xmark  & \xmark  & \cmark      \\
    T-LESS \cite{Hodan2017tless}          & pose estimation       & real        & 30                & 30k       & 20        & 3               & \cmark& \cmark     & \xmark       & \xmark  & \xmark  & \cmark      \\
    SUN RGB-D \cite{Song2015}             & semantic segmentation & real        & 800               & 10k       & 47        & 4               & \cmark& \xmark     & \cmark$^2$   & \cmark  & \cmark  & \xmark      \\
    DROT \cite{Rotman2016drot}            & depth reconstruction  & real        & 3                 & 0.1k      & 5         & 3               & \cmark& \cmark     & \cmark       & \xmark  & \xmark  & \cmark      \\ \bottomrule
  \end{tabularx}

  $^1$LINEMOD only provides the 6DoF pose for a single object per scene.

  $^2$SUN RGB-D does not provide per-pixel labeling; instead, image regions are more coarsely labeled using a 2D polygon.
\end{table*}

The YCB object and model set \cite{Calli2015icar,Calli2017ijrr} is a set of 77
physical everyday objects intended to be a standard benchmark for grasping and
manipulation research, available to research groups worldwide. The set of
physical objects is accompanied by a set of 103 high-resolution textured
meshes; the number of meshes differs from the number of objects, because meshes
are not available for all objects, while some objects consist of multiple
parts (with one mesh each).

A selection of 21 YCB objects was used in the YCB-Video dataset
\cite{Xiang2017corr}. It contains 92 scenes of a handheld ASUS Xtion camera in
so-called ``fast-cropping mode'' (i.e., the images are cropped to the center
region). The FAT dataset \cite{Tremblay2018fat} is a photo-realistic synthetic
dataset that uses the same objects as the YCB-Video dataset. For comparability
with the YCB-Video and FAT datasets, we use the same objects in our YCB-M
dataset (see Sec.~\ref{sec:data_acquisition}).

LINEMOD \cite{Hinterstoisser2012accv} is a popular benchmark dataset for pose
estimation. In contrast to our dataset, however, the 6DoF ground truth pose is
only available for one object per scene.

None of the datasets presented so far has data from more than one camera. A
recent comprehensive review of 102 publicly available RGB-D datasets
\cite{Firman2016, Firman2018} lists only 4 datasets that contain data from more
than one 3D sensor per scene. Out of these, only 3 have ground truth
annotations for objects: DROT \cite{Rotman2016drot}, SUN RGB-D \cite{Song2015}
and T-LESS \cite{Hodan2017tless}. The intended use case of DROT is not pose
estimation, but depth reconstruction. As such, it contains high-quality depth
data, but only 100 frames. SUN RGB-D is intended for semantic segmentation, and
it lacks mesh models and only has 3D bounding box annotations for objects, not
the full 6DoF pose. The dataset that is most similar to ours is T-LESS: It has
data from two different RGB-D cameras and a high-resolution 2D camera, taken
from cluttered scenes of 30 different (texture-less) objects.

\subsection{Comparisons of 3D cameras}

There are several works performing comparative evaluations of 3D cameras.
Halmetschlager-Funek et al.~\cite{HalmetschlagerFunek2018ram} perform a
comparative evaluation of ten 3D cameras (four of which are also present in our
dataset) with regards to sensor noise and bias, sensitivity to target material
properties and illumination, and sensor interference.
Several other works report similar experiments for the Kinect v1 (structured
light) and v2 (ToF) sensors \cite{Wasenmueller2016comparison,
Sarbolandi2015kinect1vs2, Jing2017comparison}.

These papers are complementary to our present work: While these approaches
quantify sensor properties such as noise, depth precision, and temperature
drift, the test scenes are necessarily artificial (e.g., the sensor pointing at
a uniform flat surface). Our dataset provides real-world measurements from the
sensors and offers the opportunity of directly testing algorithms against each
sensor.

\section{The YCB-M Dataset}
\label{sec:dataset}

This section presents the methods used for data acquisition and ground truth annotation of our dataset. It also gives an overview of the used cameras.
The dataset is available under the following URL: \url{https://doi.org/10.5281/zenodo.2579172}.

\subsection{Data Acquisition}
\label{sec:data_acquisition}

For our dataset, we recorded 32 scenes containing a subset of the YCB Object and Model Set \cite{Calli2015icar}. We selected the same subset of objects used in the YCB-Video \cite{Xiang2017corr} and FAT \cite{Tremblay2018fat} datasets, except for the ``Master Chef Can'', which was missing from our copy of the object set.
Each scene contains between 3 and 8 objects (5 on average) with varying degrees of occlusion, and each object occurs on average in eight different scenes throughout the dataset.
Additionally, a marker board containing markers for camera pose estimation is present in every scene (see Sec.~\ref{sec:ground_truth_annotation}).

\begin{figure*}[htbp]
  \centering
  \includegraphics[height=6cm]{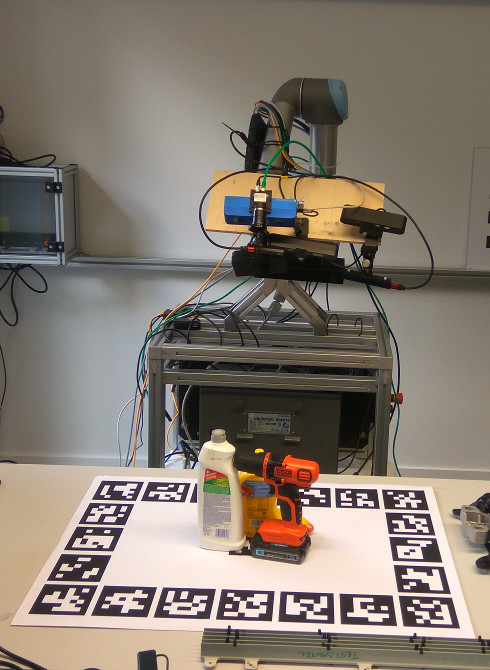}
  \quad
  \includegraphics[height=6cm]{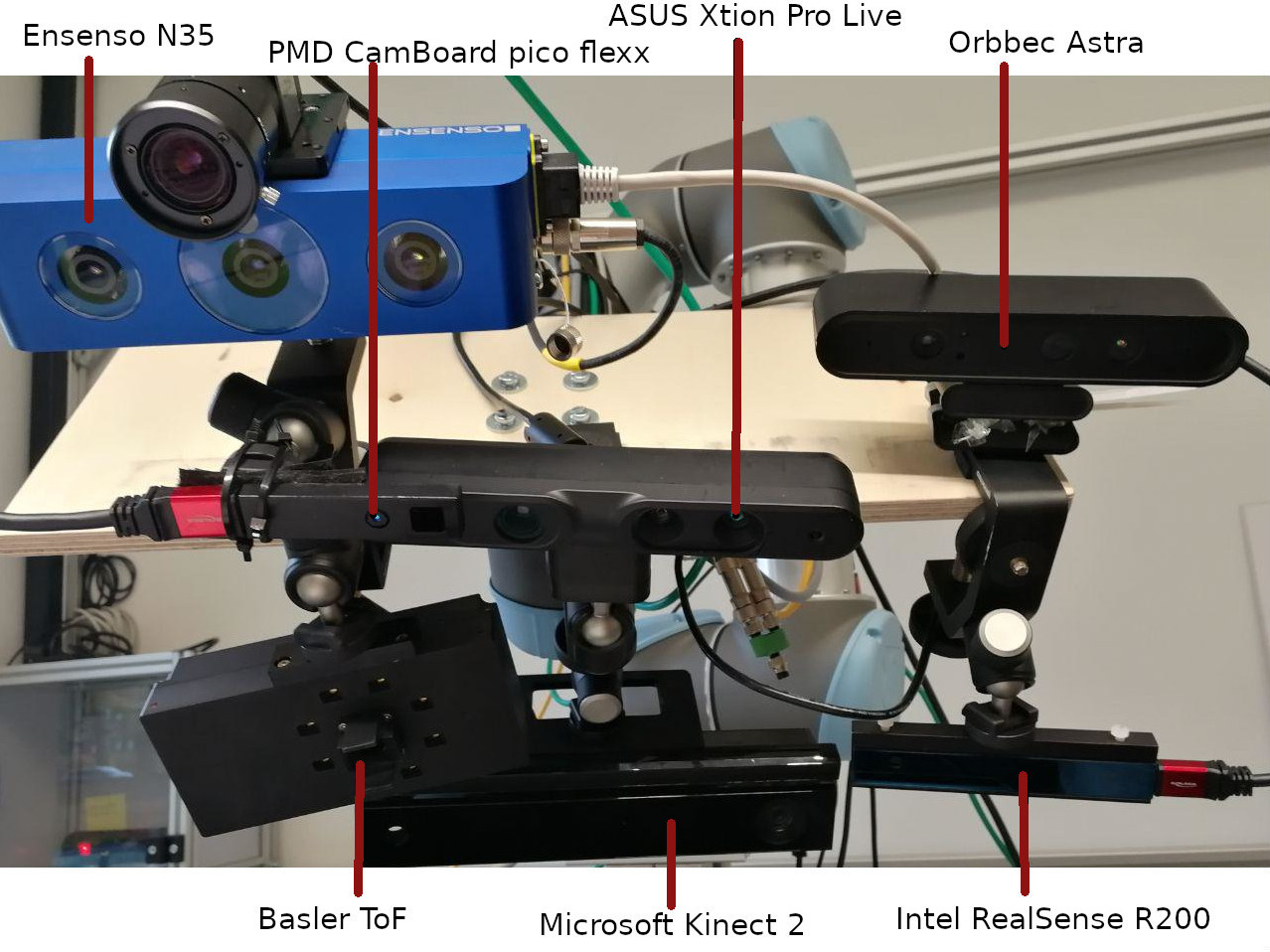}
  \caption{Left: The robot arm with the camera rig while capturing a scene. Right: Close-up of the capturing rig with all seven tested cameras.}
  \label{fig:multicam_rig}
\end{figure*}

To provide data from different angles while leaving the transformation between the cameras constant, we mounted the cameras to the end effector of a UR-5 robot arm (see Fig.~\ref{fig:multicam_rig}). The robot arm followed a fixed trajectory with 9 anchor points.

Ideally, the data from all cameras would have been recorded simultaneously. However, this would lead to degraded depth data due to interference between the cameras, since most of the cameras operate in the same infrared spectrum. For this reason, we recorded the data in two phases while ensuring that only one camera is active at a given time. In the first phase (labeled ``snapshots'' in the dataset) the arm would stop at each of the anchor points and record one frame from each camera in turn. In the second phase (labeled ``trajectory'' in the dataset), the robot arm would move through all anchor points without stopping while recording with a single camera (again, to avoid interference). Thus the arm would perform the motion once in the first phase and 7 times (once for each camera) in the second phase.

\subsection{Ground Truth Annotation}
\label{sec:ground_truth_annotation}

In addition to the raw data consisting of RGB (resp.\ grayscale) images and depth images, we provide ground truth annotations for every frame. Primarily we provide the 6DoF pose of each object present in the frame, as well as per-pixel segmentation images (see first two rows of Fig.~\ref{fig:gt_seg_dope}), 2D and 3D bounding boxes and the camera position relative to the global reference frame (the marker in the center of the marker board). Additionally, we include high-resolution meshes (from the YCB model set \cite{Calli2015icar}) of the objects used in our dataset.
The annotation follows the format of the synthetic FAT dataset \cite{Tremblay2018fat}, encouraging reusability of the format and complementing existing work. Our dataset can directly be used to further train and evaluate object recognition algorithms that were trained or tested on the FAT dataset, such as DOPE~\cite{Tremblay2018dope}. Moreover, the provided annotation data makes the dataset suitable for tasks other than 6DoF pose estimation, such as object reconstruction, semantic segmentation or 2D object detection.

The ground truth annotation was obtained by manually labeling the 6DoF poses of all objects relative to a fixed reference frame once per scene, and then transforming the poses into the camera frame for each image pair. The manual labeling process was sped up by first generating initial guesses for the object poses using PoseCNN~\cite{Xiang2017corr}. Then, the final ground truth was created by manually refining the poses, removing false positives and adding missing objects.

The fixed reference frame is defined by a board containing 21 ArUco markers \cite{GarridoJurado2014aruco}, which is included in every scene (Fig.~\ref{fig:multicam_rig} left). Using the markers we then estimated the camera poses in the reference frame and transformed the ground truth poses into the camera frame. Unfortunately, the marker-based camera pose estimation yielded inaccurate results in some cases, causing the object poses to be inaccurately transformed into the camera frame. For this reason, we refined the marker-based camera pose estimation by matching a synthetic point cloud consisting of the true constellation of models and table plane into the data recorded by the cameras. We aim to improve the annotation further by using additional information we recorded during the process (arm joint angles and separate calibration scenes containing a checkerboard calibration pattern).

From the 6DoF object poses in the camera frame and the size of the object, the 3D bounding box can be trivially computed. The remaining annotation data (per-pixel segmentation images, 2D bounding boxes and visibility) was computed by rendering a synthetic depth image of the object models at their ground truth poses from the perspective of the camera. The segmentation mask was directly obtained from the unoccluded pixels belonging to an object in the synthetic depth image of the full scene. Slight misalignments of the ground truth pose can lead to parts of the background or other objects to be included in the segmentation mask. To compensate for this, pixels whose synthetic depth value differed by more than a certain threshold (\unit[0.04]{m}) from the actual depth image recorded by the camera were excluded from the segmentation image. The 2D bounding boxes are computed from synthetic depth images of each object rendered in isolation (thereby removing all occlusions).

The visibility of each object (a measure of the amount of occlusion) is defined as follows. We first compute the raw visibility $v_{raw}$ as:

\[v_{raw} = p_{visible} / p_{total}\]

where $p_{total}$ is the number of pixels that the object occupies in a
synthetic depth image of the object rendered in isolation, and $p_{visible}$
is the number of pixels that the object occupies in the final segmentation
image, taking occlusions from other objects into account. However, $v_{raw}$ does not
account for objects that are partially outside of the camera field of view.
Therefore, we adjust $v_{raw}$ as follows to obtain the final visibility $v$:

\[v = v_{raw} \cdot (A_{visible}  / A_{total})\]

given by $A_{visible}$, the area of the projected 3D bounding box that is
inside the image borders divided by $A_{total}$, the total area of the
projected 3D bounding box. The projected 3D bounding box is the 2D polygon that
is formed by projecting the 3D bounding box into the camera's perspective.

Objects with a visibility of 0 are removed from the frame annotation; in other
words, each object that is listed in the frame annotation json file has at
least one labeled pixel in the segmentation image (and vice versa).

\subsection{3D Cameras in the Dataset}
\label{sec:cameras}

\begin{figure*}
  \setlength\tabcolsep{1.5pt} %
  \small
  \begin{tabularx}{\linewidth}{lXXXXXXX}
    & \Centering\textbf{Xtion} & \Centering\textbf{\mbox{Astra}} & \Centering\textbf{\mbox{Kinect2}} & \Centering\textbf{\mbox{Realsense} R200} & \Centering\textbf{\mbox{pico flexx}} & \Centering\textbf{Basler ToF} & \Centering\textbf{Ensenso N35}\\

  \rotatebox{90}{\begin{minipage}{1.7cm}{\begin{center}RGB / intensity\end{center}}\end{minipage}} & %
  \includegraphics[width=\linewidth]{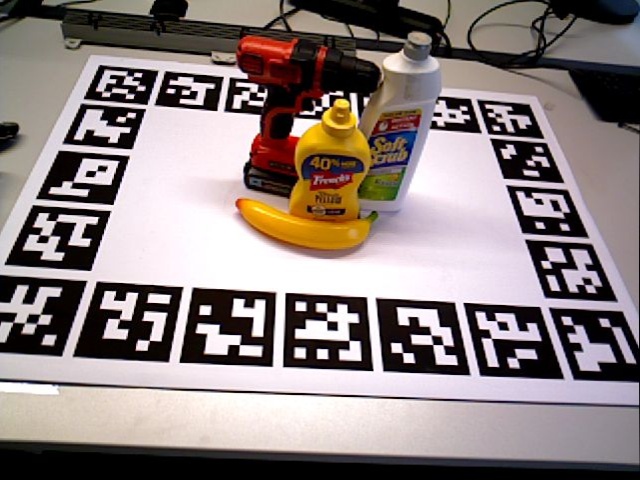} &
  \includegraphics[width=\linewidth]{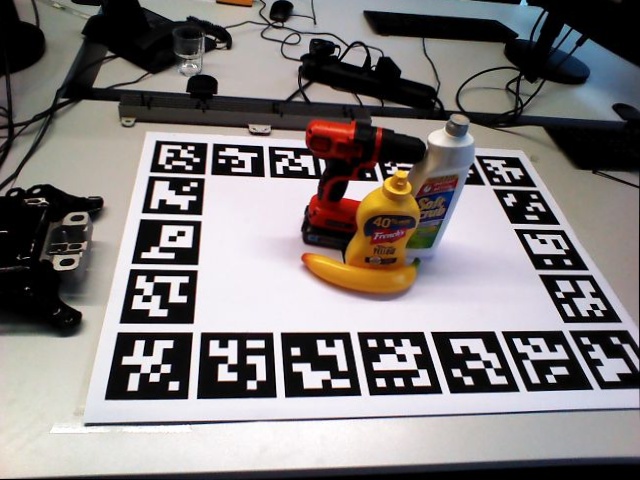} &
  \includegraphics[width=\linewidth,angle=180,origin=c]{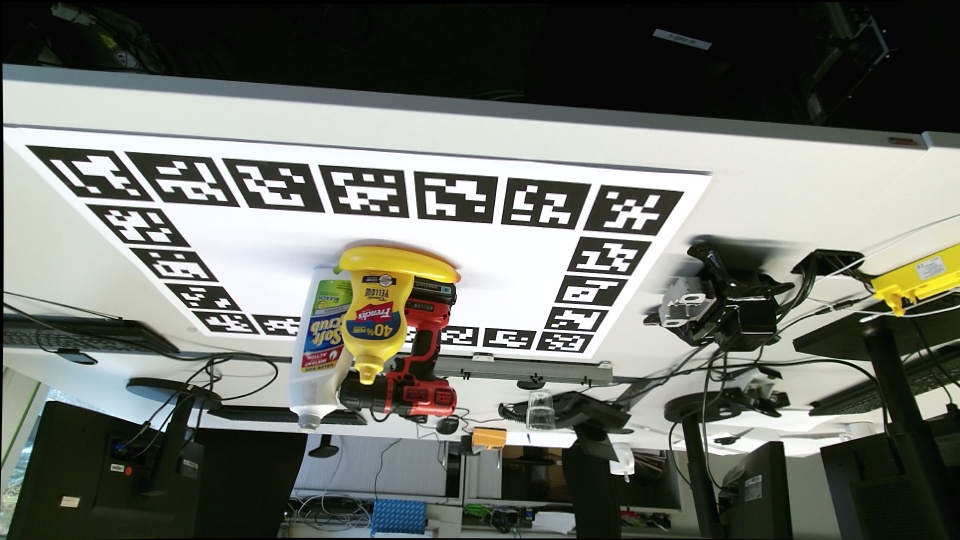} &
  \includegraphics[width=\linewidth]{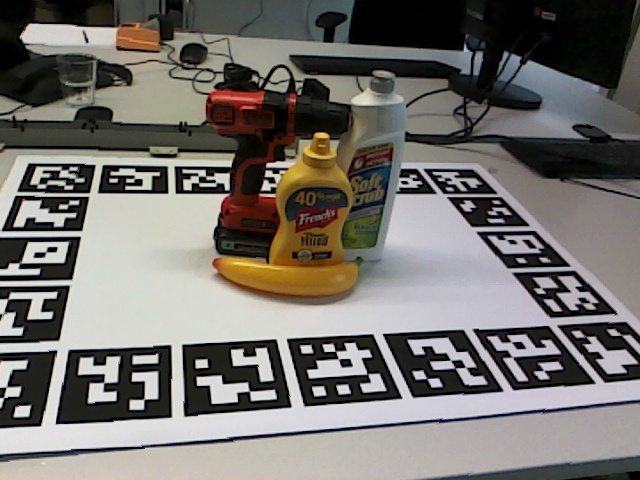} &
  \includegraphics[width=\linewidth]{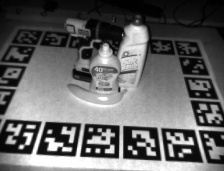} &
  \includegraphics[width=\linewidth,angle=180,origin=c]{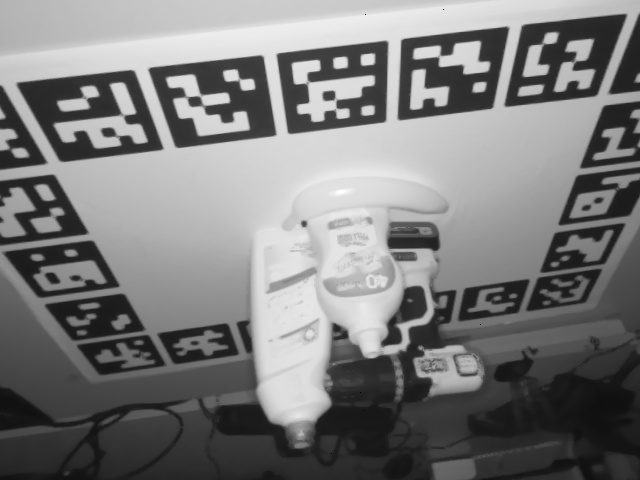} &
  \includegraphics[width=\linewidth]{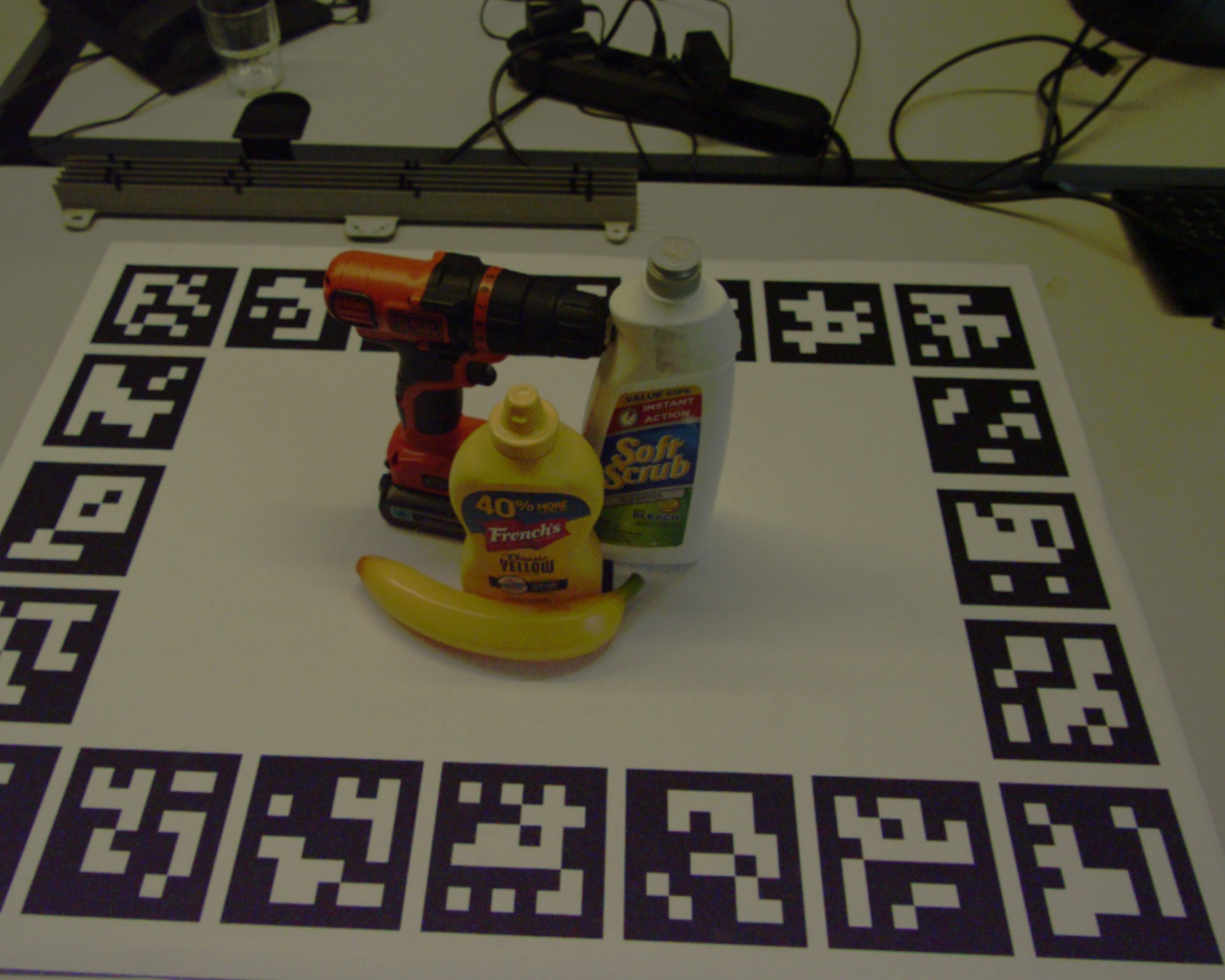}
  \\

  \rotatebox{90}{\begin{minipage}{1.7cm}{\begin{center}depth\end{center}}\end{minipage}} & %
  \includegraphics[width=\linewidth]{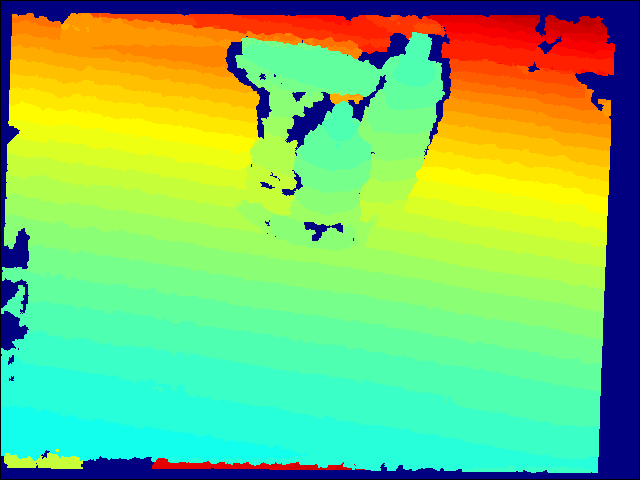} &
  \includegraphics[width=\linewidth]{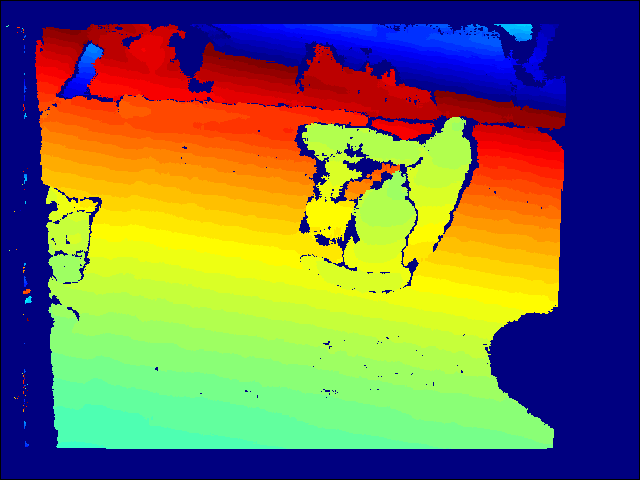} &
  \includegraphics[width=\linewidth,angle=180,origin=c]{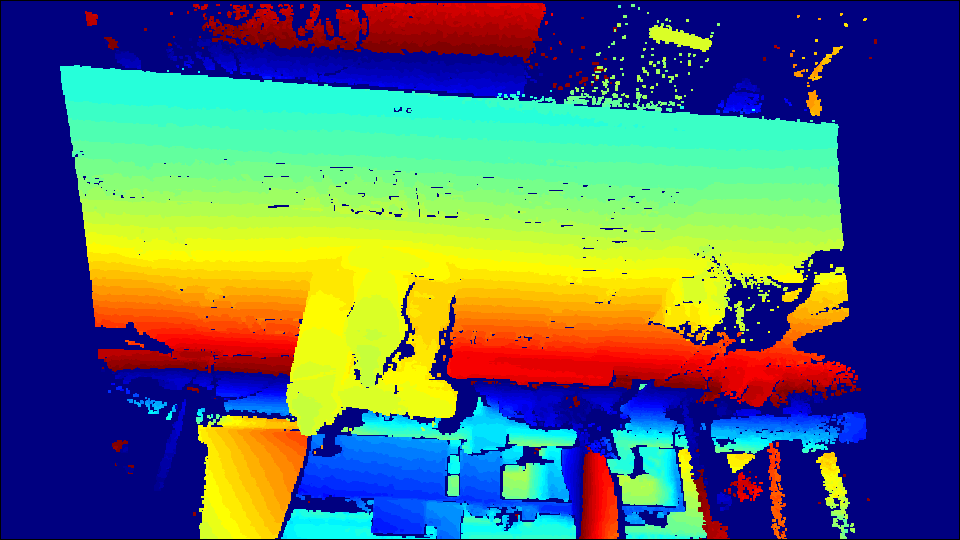} &
  \includegraphics[width=\linewidth]{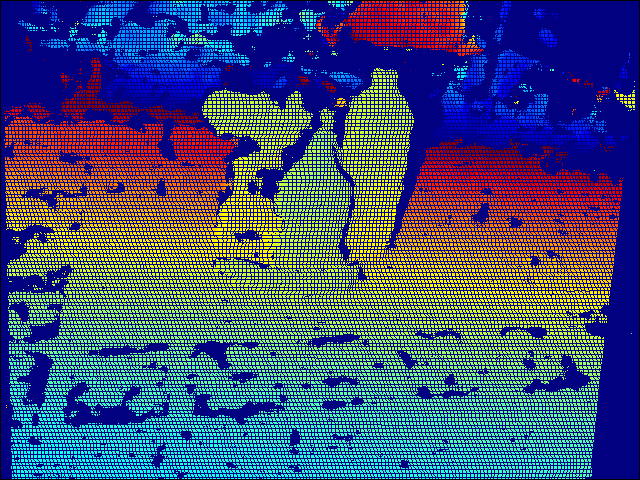} &
  \includegraphics[width=\linewidth]{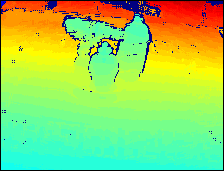} &
  \includegraphics[width=\linewidth,angle=180,origin=c]{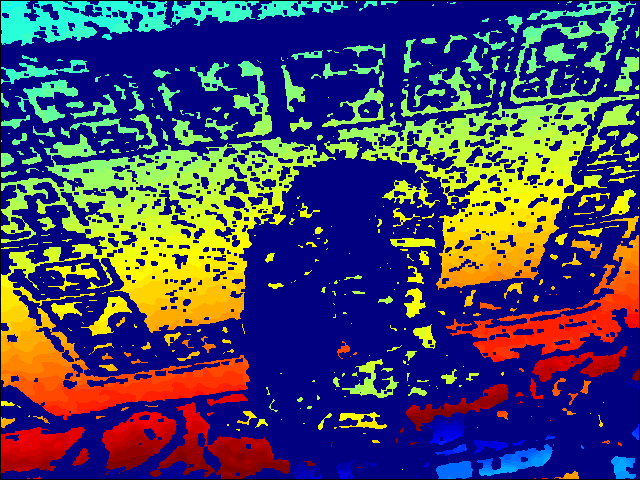} &
  \includegraphics[width=\linewidth]{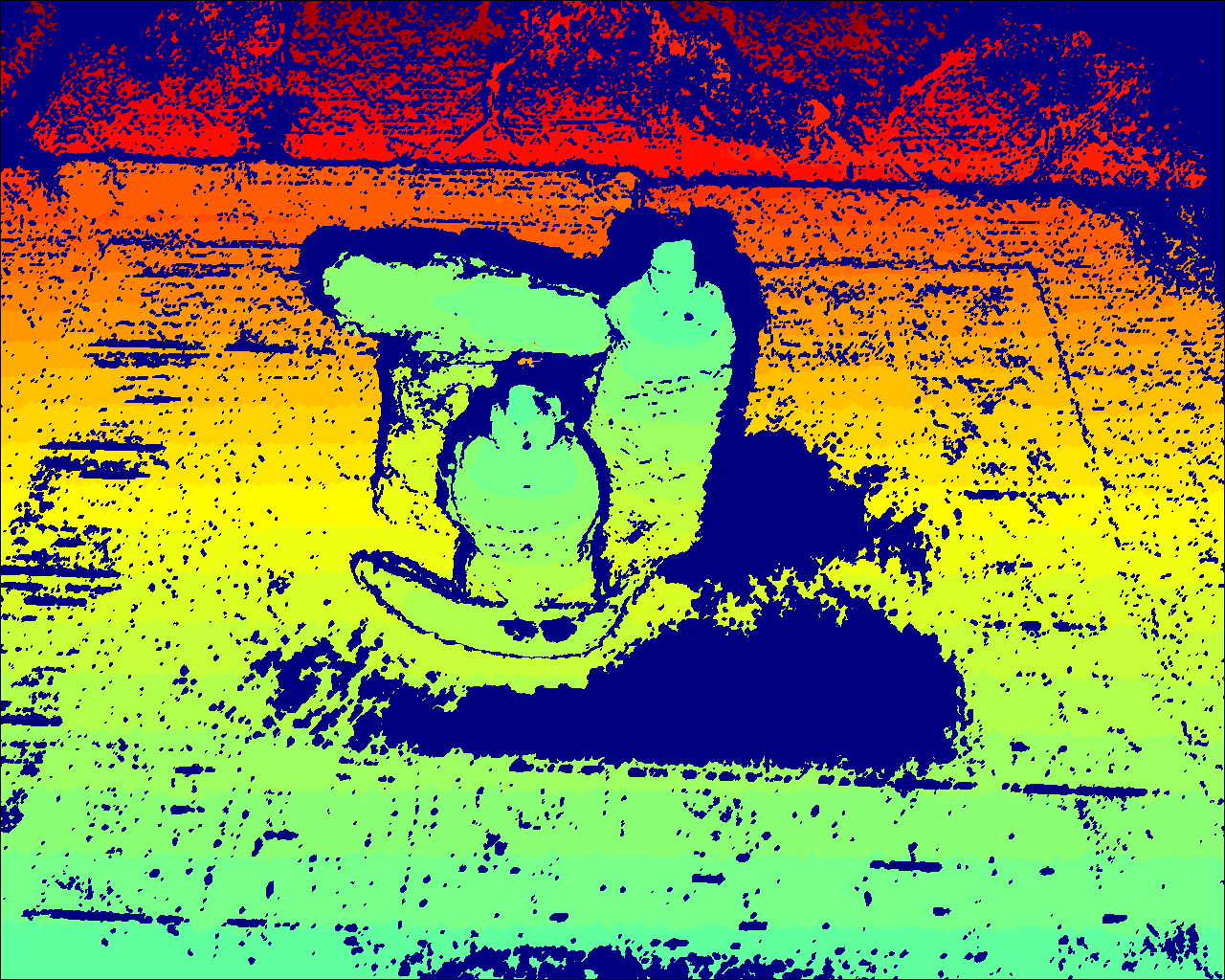}
  \\
  \rotatebox{90}{\begin{minipage}{1.7cm}{\begin{center}point cloud (full)\end{center}}\end{minipage}} & %
  \includegraphics[width=\linewidth]{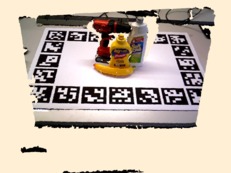} &
  \includegraphics[width=\linewidth]{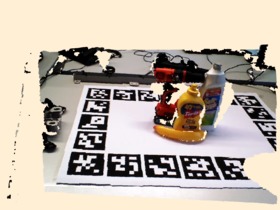} &
  \includegraphics[width=\linewidth]{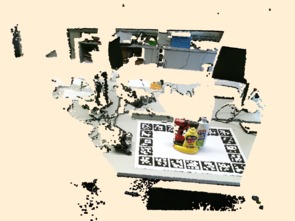} &
  \includegraphics[width=\linewidth]{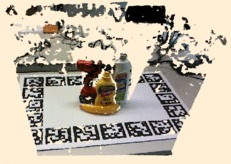} &
  \includegraphics[width=\linewidth]{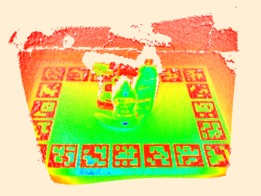} &
  \includegraphics[width=\linewidth]{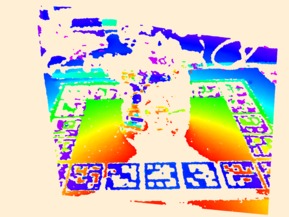} &
  \includegraphics[width=\linewidth]{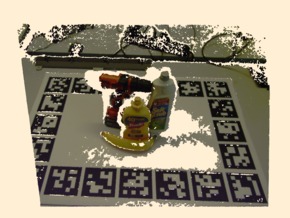}
  \\
  \rotatebox{90}{\begin{minipage}{1.7cm}{\begin{center}point cloud (detail)\end{center}}\end{minipage}} & %
  \includegraphics[width=\linewidth]{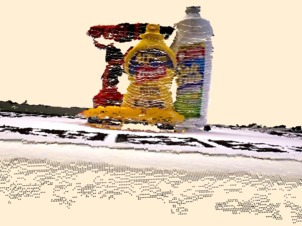} &
  \includegraphics[width=\linewidth]{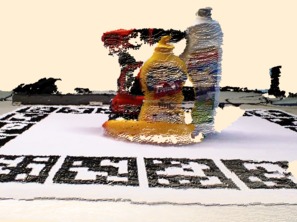} &
  \includegraphics[width=\linewidth]{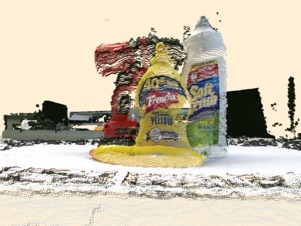} &
  \includegraphics[width=\linewidth]{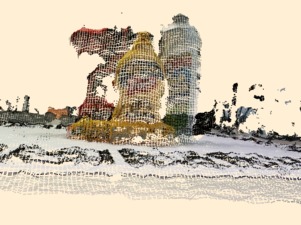} &
  \includegraphics[width=\linewidth]{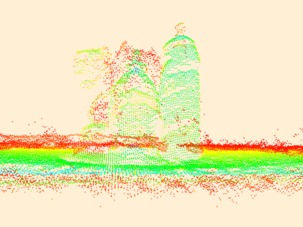} &
  \includegraphics[width=\linewidth]{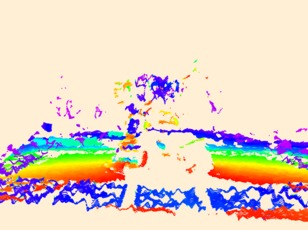} &
  \includegraphics[width=\linewidth]{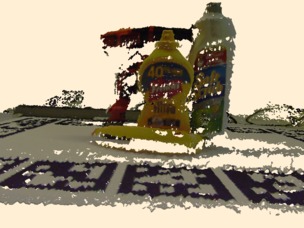}
\end{tabularx}
\caption{Camera data from all cameras recorded from the robot position shown in Fig.~\ref{fig:multicam_rig}.}
\label{fig:raw_data}
\end{figure*}

This section gives an overview of the cameras in our dataset (see
Table~\ref{tab:camera_specs}).
For illustration, Fig.~\ref{fig:raw_data} shows one frame of an example scene from
our dataset as recorded by each camera.\footnote{While our dataset only includes
the RGB/intensity and depth images, we will provide a tool to transform them
into point clouds.} The robot remained stationary while each of these frames
was recorded; the differences in perspective stem from the different mounting
positions of the cameras and their different optics.

\begin{table*}
  \caption{Technical specifications of the cameras in the dataset, as provided by the manufacturers.}
  \label{tab:camera_specs}

  \begin{center}
    \small
    \begin{tabularx}{\textwidth}{lXXXXXXX}
      \toprule
      & \Centering\textbf{ASUS Xtion} & \Centering\textbf{Orbbec \mbox{Astra}} & \Centering\textbf{Microsoft \mbox{Kinect2}} & \Centering\textbf{Intel \mbox{Realsense} R200} & \Centering\textbf{pmd CamBoard \mbox{pico flexx}} & \Centering\textbf{Basler ToF} & \Centering\textbf{Ensenso \mbox{N35-604-16-bl} + RGB}\\
      \midrule
     Technology        & Struct.~Light & Struct.~Light & ToF & Act.~Stereo & ToF & ToF & Act.~Stereo \\
     Depth Resolution  & $640 \times 480$   & $640 \times 480$    & $512 \times 424$   & $628 \times 468$        & $224 \times 171$         & $640 \times 480$       & $1280 \times 1024$     \\
     RGB Resolution    & $640 \times 480$   & $640 \times 480$    & \mbox{$1920 \times 1080^*$} & \mbox{$1920 \times 1080^*$}      & ---               & ---             & $1280 \times 1024$     \\
     Intensity Resolution & ---         & ---          & ---             & ---                  & $224 \times 171$         & $640 \times 480$       & ---             \\
     Range [m]         & 0.8--3.5    & 0.6--8.0     & 0.5--4.5        & 0.5--4.0             & 0.1--4.0          & 0.5--6.6        & 0.33--1.1       \\
     Opening Angle [$^\circ$]  & $58 \times 45$     & $60 \times 49.5$    & $70.6 \times 60$       & $59 \times 46$              & $62 \times 45$           & $57 \times 43$         & $62 \times 48$         \\  %
     Frames per Second & 30          & 30           & 30              & 30--60               & 5--45             & 15              & 10              \\
     Housing           & ---         & ---          & ---             & ---                  & ---               & IP30            & IP65 / IP67     \\
     Weight [g]        & 230         & 250          & 1,400           & 65                   & 8                 & 400             & 650             \\
     Power Consumption [W] & $<2.5$  & $<2.4$       & avg.~16         & $<1.6$               & $<0.5$            & $<15$           & ?               \\
     Interface         & USB 2       & USB 2        & USB 3           & USB 3                & USB 3             & Ethernet        & Ethernet        \\
     \bottomrule
    \end{tabularx}
  \end{center}

  $^*$In the dataset, Kinect2 images were captured at
  $960 \times 540$ pixels and Realsense R200 images at $640 \times 480$
  pixels. All other cameras were captured at the maximum resolution given
  in the table.
\end{table*}

The cameras selected for the dataset cover all three major RGB-D technologies
(structured light, Time-of-Flight and (active) stereo), as well as both
consumer and industrial cameras (both of which are commonly used in robotics
research).

The ASUS Xtion Pro Live belongs to the family of Primesense-based structured
light cameras (along with the Kinect 1, the Primesense Carmine and the Occipital
Structure Sensor) and has long been the most popular 3D camera used in
robotics. Compared to many of the other cameras, the depth image typically has
low noise and a low number of missing values.

After the Xtion was discontinued, Orbbec brought its Astra camera to market,
and it has become a popular drop-in replacement for the Xtion, since the
specifications, software API and camera data are almost identical to the Xtion.
The depth image quality is also almost identical but tends to have a small
number of artifacts at specific depth ranges and near depth discontinuities and
image borders (this is visible near the left depth image border in
Fig.~\ref{fig:raw_data}). Since these artifacts are predictable, they can be
filtered out in a postprocessing step.

The ToF-based Kinect2 has the largest field of view of the tested cameras. Its
depth data is less smooth than the structured light sensors, but for ranges
greater than \unit[2]{m}, it is more precise \cite{HalmetschlagerFunek2018ram}.

The Intel Realsense R200 camera is a small and light-weight infrared active
stereo camera with an additional RGB camera. Its RGB camera has the highest
resolution (along with the Kinect2) of all tested cameras. The depth data is
not as smooth as the structured light cameras (e.g., compare the planarity of
the table visible in the point clouds in Fig.~\ref{fig:raw_data}). Moreover, in
some frames, failures of the stereo matching lead to a high amount of
artifacts, which can be problematic for applications sensitive to misreported
depth values (such as object modeling or obstacle avoidance).

The pmd CamBoard pico flexx is a very small ToF camera. Its absence of an RGB
camera and its low resolution make it not well-suited for some tasks such as
mid-range object recognition; however, its low minimum range and small size
make it attractive for other tasks such as in-hand object pose estimation. The
precision of the depth data is roughly comparable to the Kinect2.

The Basler ToF ES camera also lacks a dedicated RGB camera. Its depth image
typically has many missing values and the highest noise of all tested cameras.

The Ensenso N35 close-range active stereo camera is available with several
different optics and either an infrared or visible blue light projector. The
tested model (N35-604-16-bl) has the second-to-largest operating range and a
blue light projector as well as an additional uEye RGB camera. Within its
operating range, the Ensenso camera has the highest precision of all tested
cameras.

\section{Evaluation}

In order to demonstrate the usability of our dataset, we evaluate the performance of the DOPE \cite{Tremblay2018dope} pose recognition algorithm on the different cameras. Of course, the opposite (evaluating the performance of a given camera on different algorithms), or assessing a combination between different camera models and algorithms would also be possible.
We first show some example results from the DOPE object recognition
before demonstrating a quantitative evaluation based on our data set.

\begin{figure*}
  \setlength\tabcolsep{1.5pt} %
  \begin{tabularx}{\linewidth}{l>{\centering\arraybackslash}X>{\centering\arraybackslash}X>{\centering\arraybackslash}X>{\centering\arraybackslash}X>{\centering\arraybackslash}X>{\centering\arraybackslash}X>{\centering\arraybackslash}X}

  \scriptsize\rotatebox{90}{\hspace{0.45cm}\textbf{GT bboxes}\hspace{0.45cm}} &
  \includegraphics[width=\linewidth]{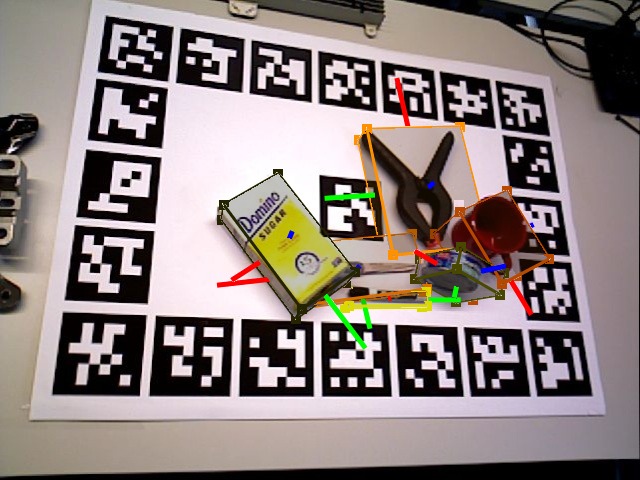} &
  \includegraphics[width=\linewidth]{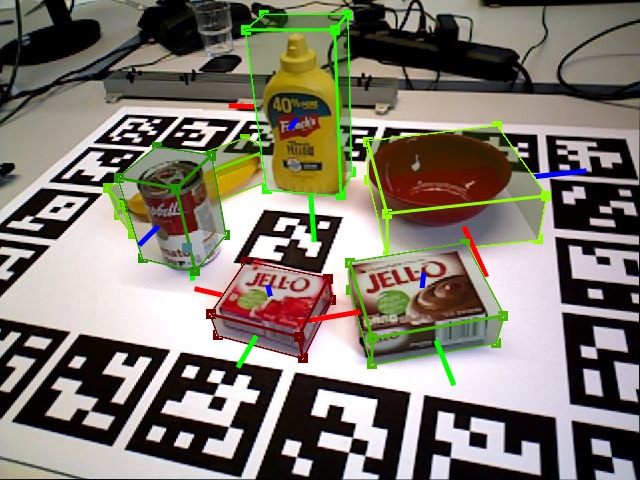} &
  \includegraphics[width=\linewidth]{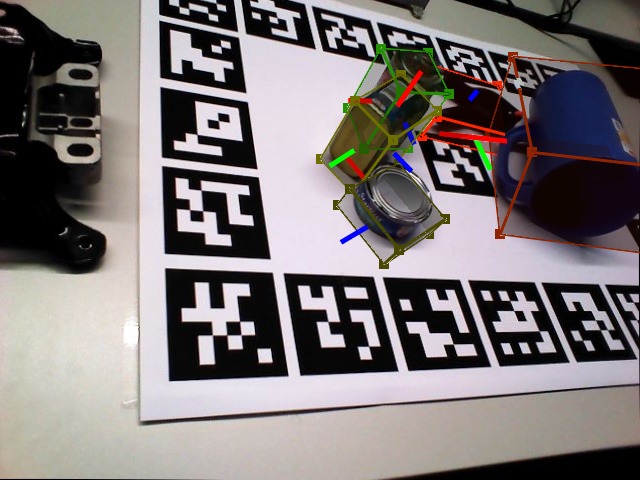} &
  \includegraphics[width=\linewidth]{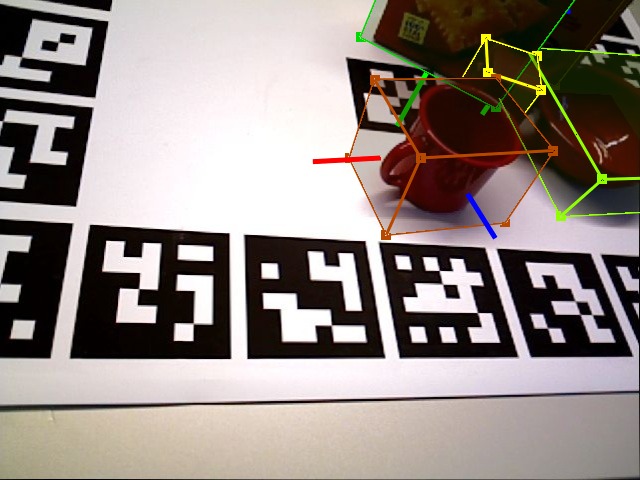} &
  \includegraphics[width=\linewidth]{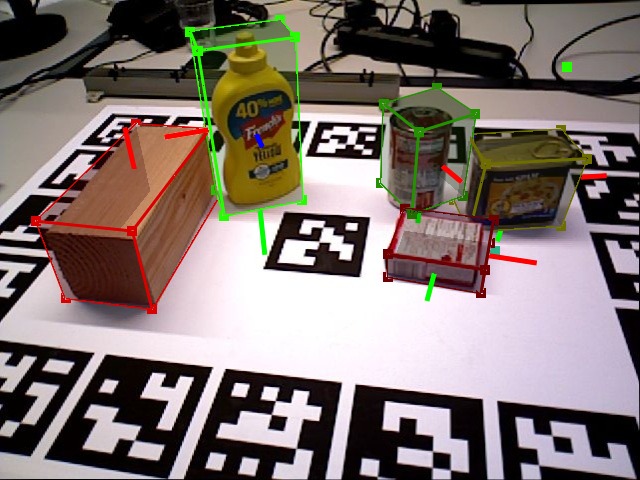} &
  \includegraphics[width=\linewidth]{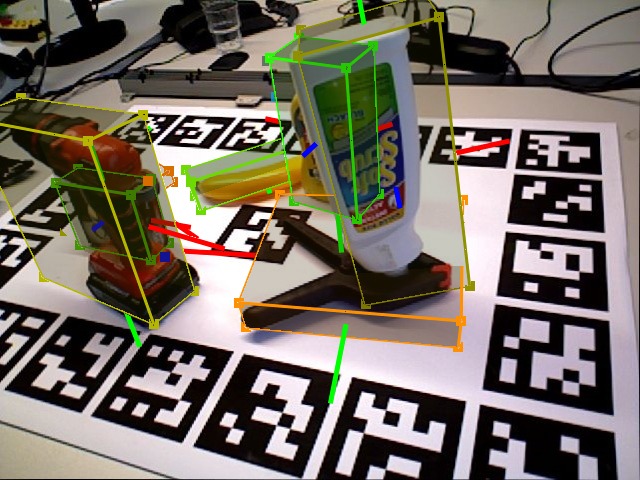}
  \\

  \scriptsize\rotatebox{90}{\hspace{0.12cm}\textbf{GT segmentation}\hspace{0.12cm}} &
  \includegraphics[width=\linewidth]{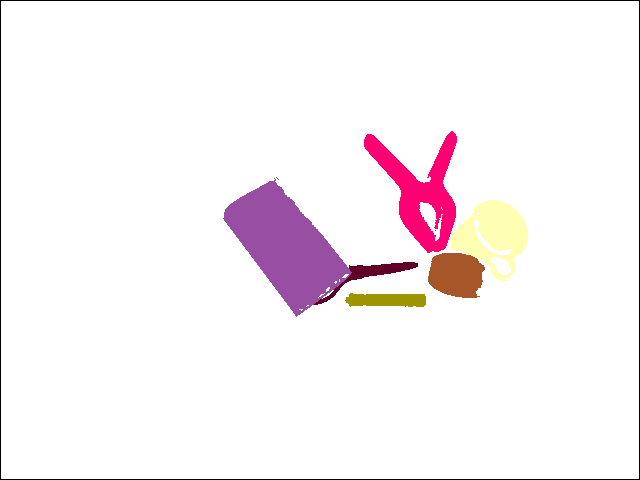} &
  \includegraphics[width=\linewidth]{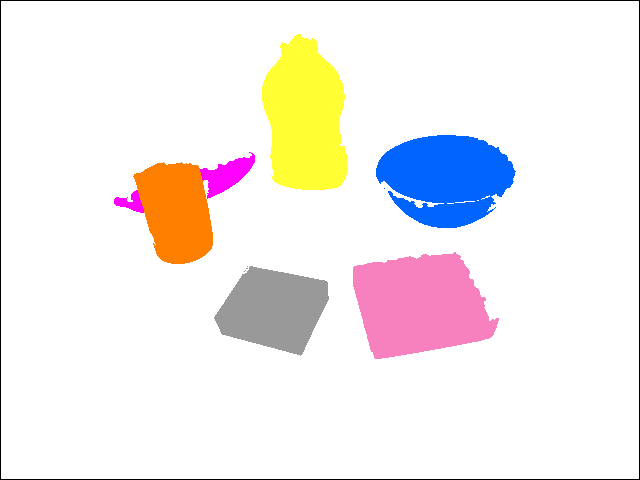} &
  \includegraphics[width=\linewidth]{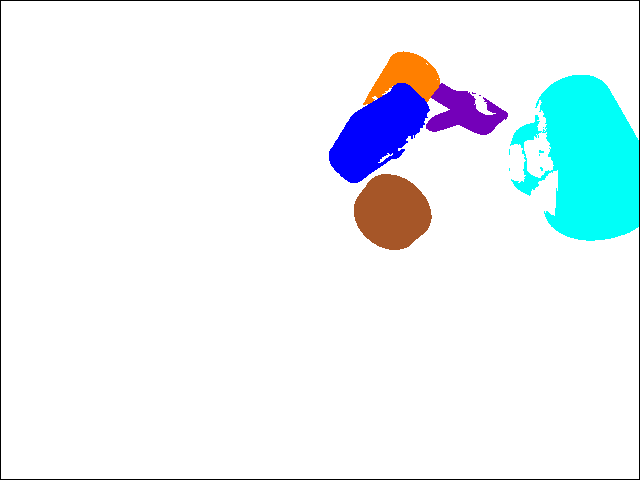} &
  \includegraphics[width=\linewidth]{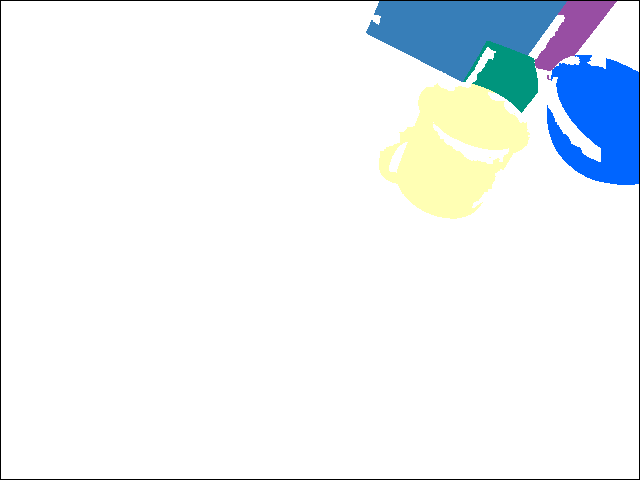} &
  \includegraphics[width=\linewidth]{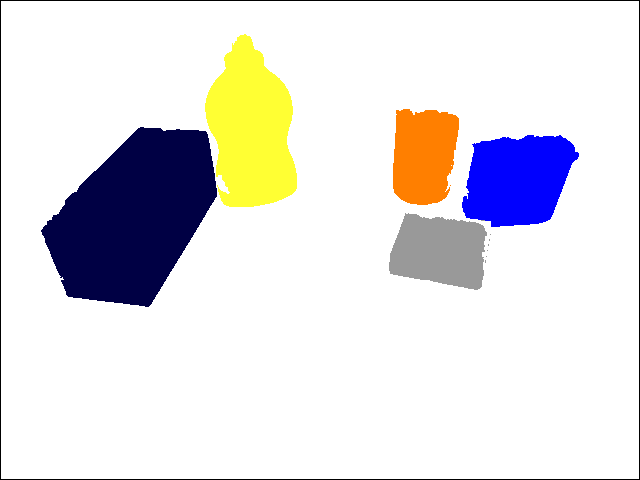} &
  \includegraphics[width=\linewidth]{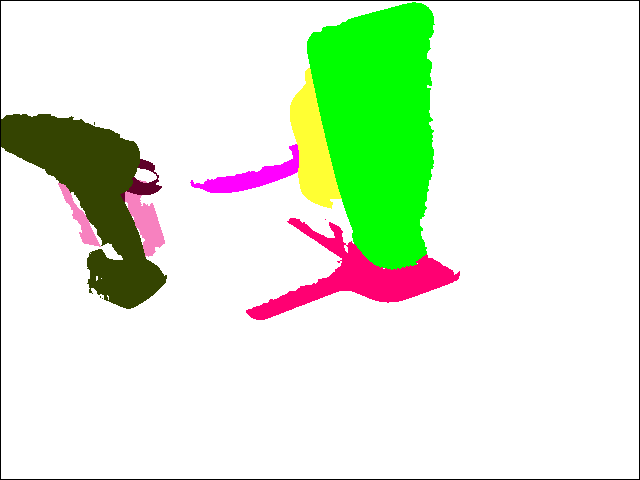}
  \\

  \scriptsize\rotatebox{90}{\hspace{0.3cm}\textbf{DOPE results}\hspace{0.3cm}} &
  \includegraphics[width=\linewidth]{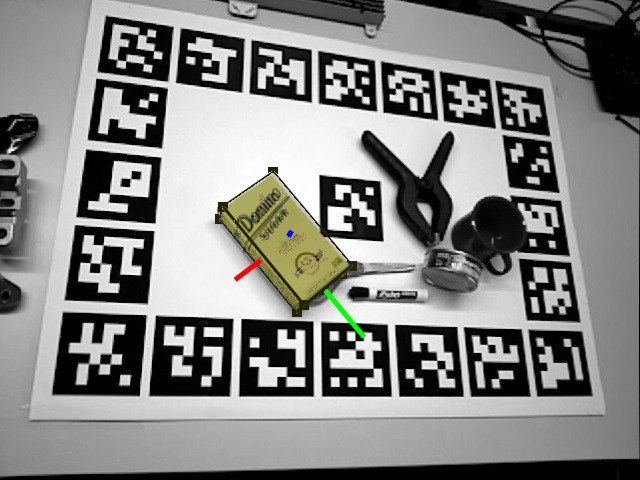} &
  \includegraphics[width=\linewidth]{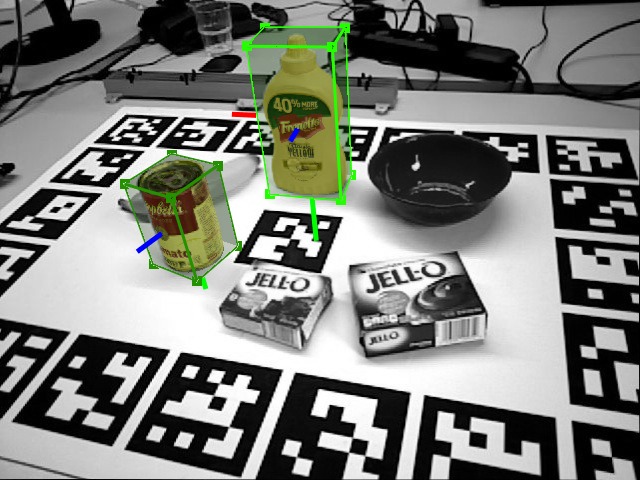} &
  \includegraphics[width=\linewidth]{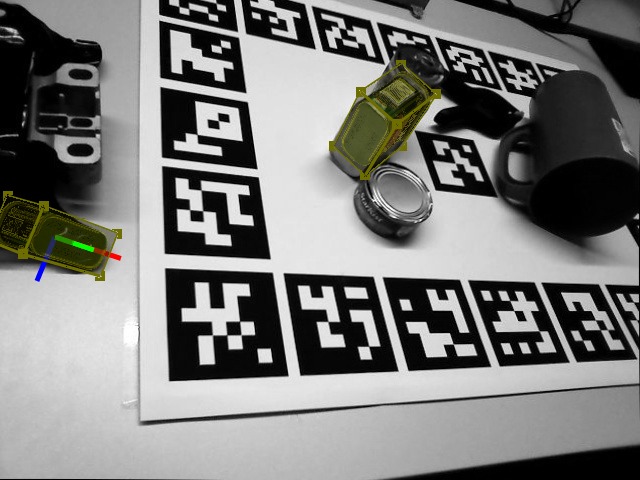} &
  \includegraphics[width=\linewidth]{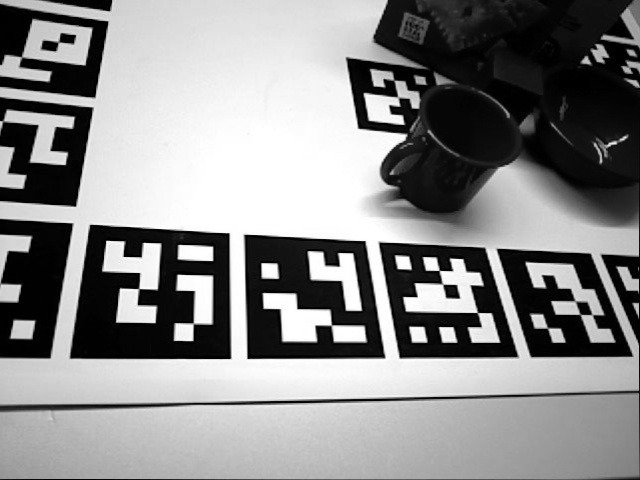} &
  \includegraphics[width=\linewidth]{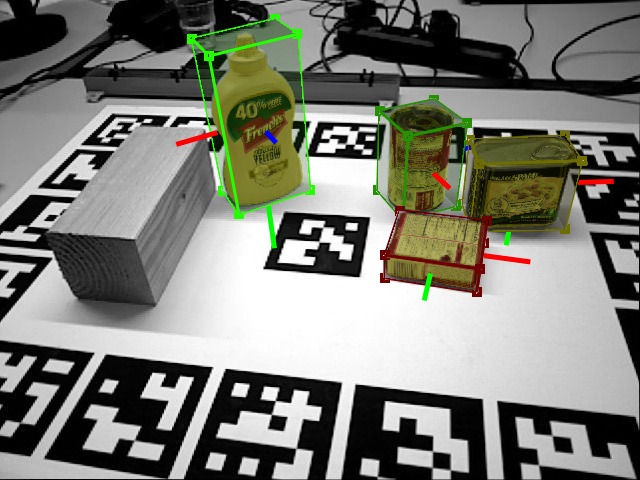} &
  \includegraphics[width=\linewidth]{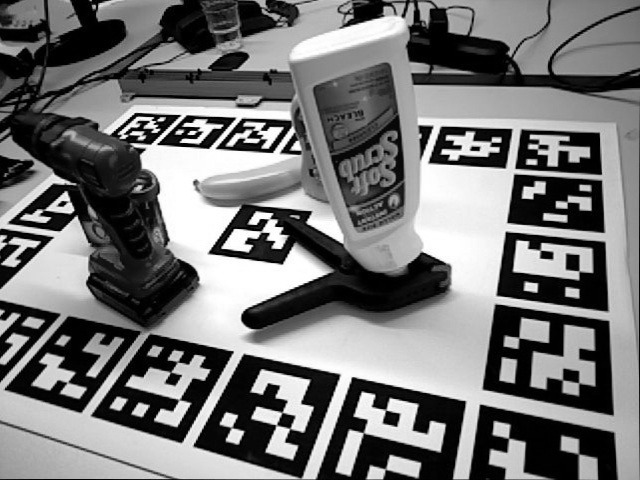}
  \\
  &
  \scriptsize (1 TP / 0 FP / 0 FN) &
  \scriptsize (2 TP / 0 FP / 0 FN) &
  \scriptsize (1 TP / 1 FP / 1 FN) &
  \scriptsize (0 TP / 0 FP / 2 FN) &
  \scriptsize (4 TP / 0 FP / 0 FN) &
  \scriptsize (0 TP / 0 FP / 1 FN)
\end{tabularx}
{%
  \definecolor{color002}{RGB}{228, 26 , 28 }
  \definecolor{color003}{RGB}{55 , 126, 184}
  \definecolor{color004}{RGB}{152, 78 , 163}
  \definecolor{color005}{RGB}{255, 127, 0  }
  \definecolor{color006}{RGB}{255, 255, 51 }
  \definecolor{color007}{RGB}{166, 86 , 40 }
  \definecolor{color008}{RGB}{247, 129, 191}
  \definecolor{color009}{RGB}{153, 153, 153}
  \definecolor{color010}{RGB}{0  , 0  , 255}
  \definecolor{color011}{RGB}{255, 0  , 255}
  \definecolor{color019}{RGB}{0  , 255, 248}
  \definecolor{color021}{RGB}{0  , 255, 0  }
  \definecolor{color024}{RGB}{0  , 101, 255}
  \definecolor{color025}{RGB}{255, 255, 180}
  \definecolor{color035}{RGB}{52 , 68 , 1  }
  \definecolor{color036}{RGB}{0  , 0  , 68 }
  \definecolor{color037}{RGB}{96 , 0  , 41 }
  \definecolor{color040}{RGB}{158, 147, 0  }
  \definecolor{color051}{RGB}{116, 0  , 185}
  \definecolor{color052}{RGB}{255, 0  , 114}
  \definecolor{color061}{RGB}{0  , 149, 125}

  \setlength{\multicolsep}{-6pt}
  \setlength{\columnsep}{0cm}
  \begin{multicols}{5}
    \scriptsize
    \protect\tikz{\draw [fill=color003] (0,0) rectangle ++(0.2,0.2);} \textbf{\texttt{003\_cracker\_box}} \\
    \protect\tikz{\draw [fill=color004] (0,0) rectangle ++(0.2,0.2);} \textbf{\texttt{004\_sugar\_box}} \\
    \protect\tikz{\draw [fill=color005] (0,0) rectangle ++(0.2,0.2);} \textbf{\texttt{005\_tomato\_soup\_can}} \\
    \protect\tikz{\draw [fill=color006] (0,0) rectangle ++(0.2,0.2);} \textbf{\texttt{006\_mustard\_bottle}} \\
    \protect\tikz{\draw [fill=color007] (0,0) rectangle ++(0.2,0.2);} \texttt{007\_tuna\_fish\_can} \\
    \protect\tikz{\draw [fill=color008] (0,0) rectangle ++(0.2,0.2);} \texttt{008\_pudding\_box} \\
    \protect\tikz{\draw [fill=color009] (0,0) rectangle ++(0.2,0.2);} \textbf{\texttt{009\_gelatin\_box}} \\
    \protect\tikz{\draw [fill=color010] (0,0) rectangle ++(0.2,0.2);} \textbf{\texttt{010\_potted\_meat\_can}} \\
    \protect\tikz{\draw [fill=color011] (0,0) rectangle ++(0.2,0.2);} \texttt{011\_banana} \\
    \protect\tikz{\draw [fill=color019] (0,0) rectangle ++(0.2,0.2);} \texttt{019\_pitcher\_base} \\
    \protect\tikz{\draw [fill=color021] (0,0) rectangle ++(0.2,0.2);} \texttt{021\_bleach\_cleanser} \\
    \protect\tikz{\draw [fill=color024] (0,0) rectangle ++(0.2,0.2);} \texttt{024\_bowl} \\
    \protect\tikz{\draw [fill=color025] (0,0) rectangle ++(0.2,0.2);} \texttt{025\_mug} \\
    \protect\tikz{\draw [fill=color035] (0,0) rectangle ++(0.2,0.2);} \texttt{035\_power\_drill} \\
    \protect\tikz{\draw [fill=color036] (0,0) rectangle ++(0.2,0.2);} \texttt{036\_wood\_block} \\
    \protect\tikz{\draw [fill=color037] (0,0) rectangle ++(0.2,0.2);} \texttt{037\_scissors} \\
    \protect\tikz{\draw [fill=color040] (0,0) rectangle ++(0.2,0.2);} \texttt{040\_large\_marker} \\
    \protect\tikz{\draw [fill=color051] (0,0) rectangle ++(0.2,0.2);} \texttt{051\_large\_clamp} \\
    \protect\tikz{\draw [fill=color052] (0,0) rectangle ++(0.2,0.2);} \texttt{052\_extra\_large\_clamp} \\
    \protect\tikz{\draw [fill=color061] (0,0) rectangle ++(0.2,0.2);} \texttt{061\_foam\_brick}
  \end{multicols}
}
\vspace{8pt}
\caption{Frames from the dataset with annotations and DOPE results. 1\textsuperscript{st} row: RGB images (Astra and Xtion cameras) with ground truth poses and 3D bboxes. 2\textsuperscript{nd} row: Ground truth segmentation. 3\textsuperscript{rd} row: Objects recognized by DOPE (shown as textured meshes); background converted to grayscale for visualization purposes. 4\textsuperscript{th} row: True Positives (TP) / False Positives (FP) / False Negatives (FN). Note: DOPE has only been trained on the six object classes marked in bold in the legend.}
\label{fig:gt_seg_dope}
\end{figure*}

Before discussing the results, it should be made clear that this initial
evaluation does not capture all aspects of the dataset, and that the presented
results are not suitable to derive strong claims about the overall quality of
the used cameras, but only show how the specific pose estimation algorithm
evaluated here performs on the different cameras.
Specifically:
\begin{inparaenum}[a)]
  \item DOPE performs object recognition solely based on RGB images, so the quality of the depth data is not evaluated;
  \item DOPE was only trained on 6 of the 20 object classes;
  \item the pico flexx and Basler ToF cameras do not provide an RGB image, so the RGB-trained DOPE models are not applicable; and finally,
  \item other aspects such as form factor or power consumption of the cameras are disregarded.
\end{inparaenum}

In order to make our results comparable to the evaluation in the original DOPE paper~\cite{Tremblay2018dope}, we followed the same evaluation strategy. In particular, we used the same pretrained models provided by the DOPE authors without modification. Since pretrained DOPE models are only available for 6 of the 20 object classes, the evaluation is limited to those classes; the remaining objects are treated as clutter. To avoid biasing the results, we chose not to finetune the models to our dataset. We included all 49,294 frames of the dataset in our evaluation.

Fig.~\ref{fig:gt_seg_dope} displays six sample frames from our dataset and the estimated poses from DOPE in order to give a qualitative impression of our dataset and the performance of the DOPE pose estimation. In the first row, our ground truth annotation with 3D bounding boxes and 6DoF poses are overlaid on RGB images of the Astra resp.\ Xtion cameras. The second row shows the ground truth per-pixel segmentation images of the objects. The last row shows meshes of the objects, as recognized by DOPE, overlaid on a grayscale version of the input image.
DOPE recognized eight out of the twelve objects (from the six classes it was trained on) in Fig.~\ref{fig:gt_seg_dope}
and estimated their pose with great precision, while reporting one false
positive. Out of the four unrecognized objects, three were either heavily
occluded or underexposed; only one (the Jell-O gelatin box in column two) was
not recognized despite not being occluded.
In column three, DOPE recognized a part of the clutter around the scene as the Potted Meat Can.
Across the whole dataset, about 31.76\,\% of the objects reported by DOPE were false positives. However, these typically did not match the scale of the actual object (i.e., either a small patch of the model was matched up with a full object from the image, or the full model was matched up with a small patch of the image).
Therefore, we expect it to be easy to filter out most of these false positives by comparison with the absolute scale of the object provided by depth cameras.

\begin{figure*}
  \small
  \begin{tabularx}{\linewidth}{XXXp{6cm}}
  \includegraphics[width=\linewidth]{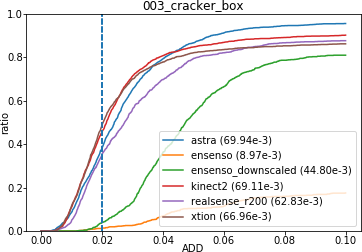} &
  \includegraphics[width=\linewidth]{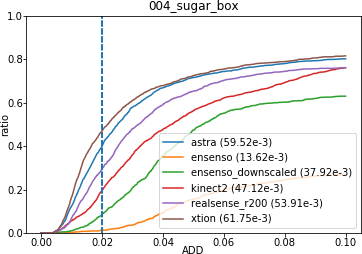} &
  \includegraphics[width=\linewidth]{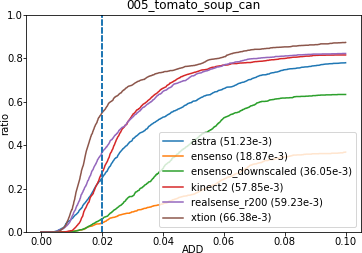} &
  \multirow{2}{*}[1.8cm]{\includegraphics[width=\linewidth]{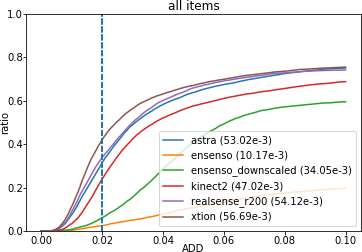}}
  \\

  \includegraphics[width=\linewidth]{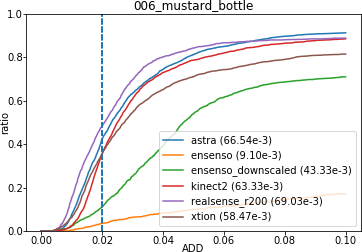} &
  \includegraphics[width=\linewidth]{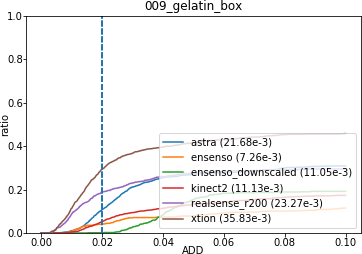} &
  \includegraphics[width=\linewidth]{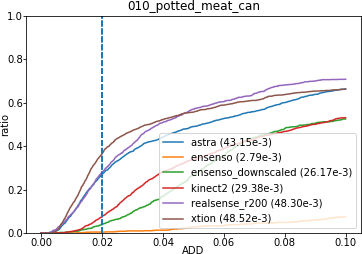} &
\end{tabularx}
\caption{DOPE recognition rates: Ratio of objects where the estimated pose is within ADD threshold from ground truth pose. The Area under the Curve (AUC) for each camera is displayed in the legends (theoretical maximum: $100 \cdot 10^{-3}$).}
\label{fig:add_plots}
\vspace{-10pt}
\end{figure*}

For the quantitative evaluation, we use the Average Distance (ADD) metric \cite{Hinterstoisser2012accv,Hodan2016poseeval} (i.e., the average distance between each model point at the ground truth pose vs.\ estimated pose) to evaluate the similarity between ground truth and estimated poses.
More precisely we evaluate the ratio of objects that were recognized within a certain margin of the ADD error (pass rate).
In each of the subfigures of Fig.~\ref{fig:add_plots}, the pass rate is plotted against the ADD required to pass, broken down by camera.
As expected, the Basler ToF and pico flexx cameras yielded a neglectable amount of true positives since they only provide a grayscale instead of an RGB image; for this reason, they were excluded from the evaluation. Furthermore, a downscaled version of the Ensenso was added, as discussed later in this section.
Additionally, we included the area under the curve (AUC) for each camera in the legend.
The dotted line represents the accuracy threshold (\unit[2]{cm}) required for grasping (adapted from \cite{Tremblay2018dope} for comparability).

The Xtion (AUC: $56.69 \cdot 10^{-3}$), Astra (AUC: $ 53.02 \cdot 10^{-3}$) and
Realsense R200 (AUC: $54.12 \cdot 10^{-3}$) cameras performed comparably well on
the dataset, while the Kinect2 performed slightly worse (AUC: $47.02 \cdot
10^{-3}$). Surprisingly, the uEye RGB camera coupled with the Ensenso performed
much worse (AUC: $10.17 \cdot 10^{-3}$). Further investigation revealed that
this was caused by the high resolution of the Ensenso image, which was not suitable for DOPE for the following reason: During
training, DOPE internally downscales all images to a height of 400 pixels, and
recognition performance drops sharply when the object appears larger (in
pixel size) in the image used for inference than in the training examples.
For this reason, we also evaluated DOPE on a version of the Ensenso frames that
were downsampled to $640 \times 512$ pixels (which is comparable to the other cameras), which drastically improved the
results (AUC: $34.05 \cdot 10^{-3}$). The remaining performance gap to the other
cameras can potentially be explained by the fact that the Ensenso RGB images in
our dataset are slightly underexposed, although DOPE is relatively robust to
extreme lighting conditions \cite{Tremblay2018dope}. The reason for this
behavior is subject to further investigation, but it demonstrates the value of
evaluating object recognition algorithms on a diverse set of cameras.

\section{Conclusions}

We have presented a large-scale training and benchmark dataset for 6DoF Pose
estimation and other related tasks that contains data from seven different 3D
cameras, spanning three major depth sensing technologies (structured light,
time-of-flight, and active stereo). We demonstrated an evaluation based on the
data set, using DOPE as a state-of-the-art RGB-based pose estimation algorithm.

In future work, we plan to apply object recognition and pose estimation approaches that make
use of the full RGB-D data (such as \cite{Faeulhammer2016icpr,Schwarz2017ijrr,Eitel2015}).
One other aspect of ongoing work is the further refinement of the provided
ground truth annotations using more sophisticated calibration techniques.

We have demonstrated that the YCB-M dataset makes a valuable contribution to
evaluating future 6DoF pose estimation and object recognition algorithms. You
are invited to retrieve and use it from
\url{https://doi.org/10.5281/zenodo.2579172}.

\bibliographystyle{IEEEtran}
\bibliography{papers}

\end{document}